\documentclass[runningheads]{llncs}
\usepackage{graphicx}
\usepackage{amsmath,amssymb} 
\usepackage{color}

\begin{document}

\title{Multi-view consensus CNN for 3D facial landmark placement} 
\titlerunning{MV-CNN for 3D facial landmark placement}

\author{Rasmus R. Paulsen\inst{1}\orcidID{0000-0003-0647-3215} \and
Kristine Aavild Juhl\inst{1}\orcidID{0000-0003-1434-697X} \and
Thilde Marie Haspang\inst{1} \and
Thomas Hansen\inst{2}\orcidID{0000-0001-6703-7762} \and
Melanie Ganz\inst{3,4}\orcidID{0000-0002-9120-8098} \and
Gudmundur Einarsson\inst{1}\orcidID{0000-0002-1326-2339}}
%
\index{Paulsen, Rasmus R.}
\index{Juhl, Kristine Aavild}
\index{Haspang, Thilde Marie}
\index{Hansen, Thomas}
\index{Ganz, Melanie}
\index{Einarsson, Gudmundur}

\authorrunning{R. R. Paulsen et al.} 


\institute{Department of Applied Mathematics and Computer Science\\Technical University of Denmark, 2800 Kgs. Lyngby, Denmark\\ \email{rapa@dtu.dk}, \url{http://www.compute.dtu.dk/} \and
Institute of Biological Psychiatry, Copenhagen University Hospital MHC Sct. Hans, Roskilde, Denmark \and
Neurobiology Research Unit, Rigshospitalet, Copenhagen, Denmark \and
Department of Computer Science, Copenhagen University, Denmark}

\maketitle

\begin{abstract}
The rapid increase in the availability of accurate 3D scanning devices has moved facial recognition and analysis into the 3D domain. 3D facial landmarks are often used as a simple measure of anatomy and it is crucial to have accurate algorithms for automatic landmark placement. The current state-of-the-art approaches have yet to gain from the dramatic increase in performance reported in human pose tracking and 2D facial landmark placement due to the use of deep convolutional neural networks (CNN). Development of deep learning approaches for 3D meshes has given rise to the new subfield called geometric deep learning, where one topic is the adaptation of meshes for the use of deep CNNs. In this work, we demonstrate how methods derived from geometric deep learning, namely multi-view CNNs, can be combined with recent advances in human pose tracking. The method finds 2D landmark estimates and propagates this information to 3D space, where a consensus method determines the accurate 3D face landmark position.
We utilise the method on a standard 3D face dataset and show that it outperforms current methods by a large margin. Further, we demonstrate how models trained on 3D range scans can be used to accurately place anatomical landmarks in magnetic resonance images.

\end{abstract}

\section{Introduction}
\label{sec:intro}

3D face recognition and analysis has a long history with important efforts dating back to work done in the early nineties~\cite{gordon1992face} and with lots of work published in the early 2000s~\cite{bowyer2006survey}. Initially, 3D scanning devices were expensive, complicated to use, and for laser scanning devices it required that the subject had to be still and have closed eyes for a longer period of time. However, the  availability of 3D scanning devices ranging from highly accurate clinical devices to consumer class products implemented in mobile devices has dramatically increased in the last decade. This means that human face recognition and analysis is moving from the 2D domain to the 3D domain. 3D morphometric analysis of human faces is an established research topic within human biology and medicine, where the applications range from 3D analysis of facial morphology~\cite{hammond20043dshort} to plastic surgery planning and evaluation~\cite{chang2015three}. Broadly speaking, the analysis of facial 3D morphometry is based on either a sparse set of landmarks that serves as a simple measure of facial anatomy or the analysis of a dense set of points that are aligned to the 3D faces using a template matching approach. The approach of matching a dense template can then be used to solve the task of selecting a few anatomical landmarks as seen in for example~\cite{grewe2016fully,paulsen2017creating}.

While a substantial amount of work has been published on automated 3D landmark placement, only a limited number of publications have drawn on the recent drastic improvements in human pose estimation based on deep learning. In this paper, we describe a framework for automated 3D landmarking of facial surfaces using deep learning techniques from the field of human pose estimation.

A classic approach to finding facial landmarks and face parameterisations consists of fitting a 3D statistical shape model to either one or several views of the 3D surface or directly to the surface~\cite{jourabloo2015pose,blanz1999morphable,zhu2016face}. These approaches use a learned statistical deformation model based on both geometry and texture variations that is able to synthesise faces within a learned low-dimensional manifold. By computing the residuals between the actual views and the synthesised face rendering, the optimal parameterisation of the face model can be found. In the seminal paper~\cite{blanz1999morphable} this is done in a standard penalised optimisation framework, while newer approaches cast the parameter optimisation into a deep learning framework~\cite{zhu2016face}. These methods work well if the learned model is broad enough to fit all new faces, and can successfully recover the face pose. Unusual facial expressions or pathologies might confuse the methods, since they fall outside the learned appearance manifold.

Placing landmarks purely on the basis of surface geometry is described in~\cite{gilani2015shape}, where landmarks are placed by computing the correspondence between a template face and a given face. The correspondence is computed by minimising bending energy between surface patches of the reference face and the target face. In~\cite{creusot2013machine} a machine learning approach is described where a set of local geometrical descriptors are extracted from facial scans and used to locate landmarks. The descriptors include surface curvature similar to what we propose. The concept of using local 3D shape descriptors to locate landmarks is also exploited in~\cite{perakis20133d}, where a facial landmark model is fitted to candidate locations found using curvature and local shape derivatives.

In this work, we use facial surfaces from a range scanner, where the data consist of surfaces with associated textures. We also use surfaces extracted as iso-surfaces from magnetic resonance (MR) images of the human head, where the face is only represented with its geometry. From a geometric point of view, the face is then a 2D surface embedded in 3D space and it is topologically equivalent to a disc. Recently, the analysis of this type of data using deep learning has seen a drastic increase under the term \emph{geometric deep learning}~\cite{bronstein2017geometric}, where one focus is the transformation of the representation of, for example, triangulated surfaces into a canonical representation that is suited for convolutional neural networks~\cite{goodfellow2016deep}. Spectral analysis of surfaces is one approach to the required domain adaptation as described in, for example,~\cite{boscaini2016learning}. Another approach is to use a volumetric representation where the surface is embedded in a 3D volume~\cite{qi2016volumetric} and the data is transformed into a volumetric occupancy grid.  The goal of the method described in~\cite{qi2016volumetric} is to classify entire objects into a set of predefined classes. Volumetric methods are still hampered by the drastic memory requirements for true 3D processing. In~\cite{qi2016volumetric} the volume size is restricted to 30x30x30, thus severely limiting the spatial resolution. An alternative approach is to render the surface or scene from multiple views and use a standard image based convolutional neural network (CNN) on the rendered views. This is described and analysed in detail in~\cite{qi2016volumetric,su2015multi}, the conclusion being that with the current memory limits and CNN architectures, multi-view approaches outperform volumetric methods. Multi-view approaches are conceptually similar to picking up an object and turning it around while looking (with one eye) for features or to identify the object. In~\cite{wiles2017silnet}, multi-view CNNs are used for silhouette prediction with convincing results.  However, there is rapid development in all approaches such as the voxel based method in~\cite{sedaghat2017orientation}, which can also estimate the orientation of the object in question.

\begin{figure}[htbp]
  \centering
  \includegraphics[width=.99\linewidth]{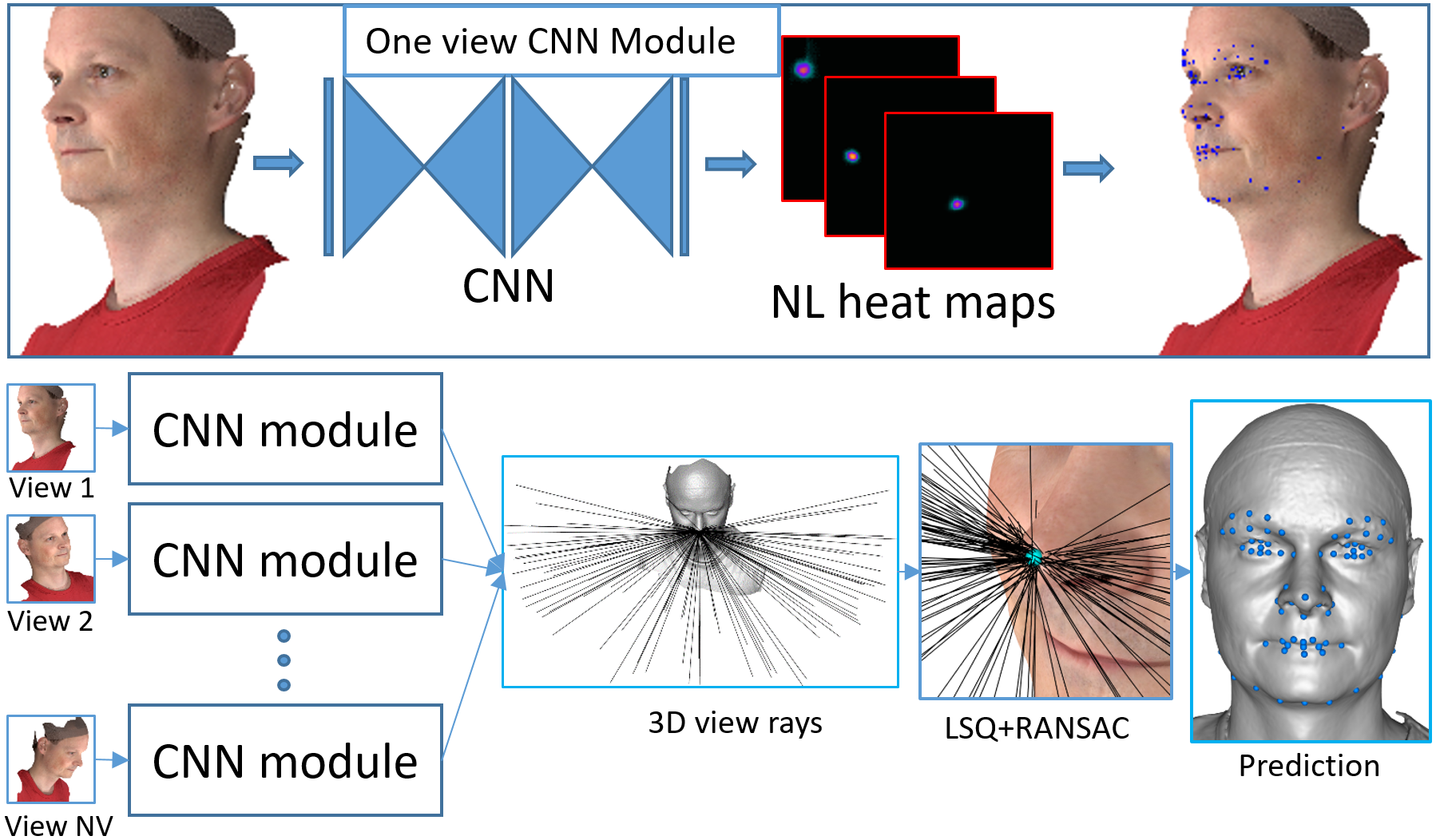}
  \caption{Overall system overview. A 3D facial scan is rendered from several views that are fed to a CNN. After a 3D view ray voting process, the result is accurately placed 3D landmarks on the face surface.}
  \label{fig:CompleteFlow}
\end{figure}

In this paper, we propose a multi-view approach to identify feature points on facial surfaces. From one direction, a feature point can be seen as a ray in 3D space. By combining results from several views, a 3D landmark can be estimated as a consensus between {\em feature rays}, similarly to classic approaches from multi-view geometry~\cite{hartley2003multiple}. A complete overview of the system can be seen in Fig.~\ref{fig:CompleteFlow}. A similar idea is presented in~\cite{Ge2016}, where a depth image from a hand is synthesised into three orthogonal views. A CNN is then trained to generate 2D heatmaps of hand landmark locations for each of these three views. The final 3D landmark locations are found by fusing the heatmap probability distributions as a set of 3D Gaussian distributions and applying a learned parameterised hand-pose subspace. We propose using more 3D projections (in this paper, 100) and an outlier-robust method to fuse the individual heatmap results. Compared to the network used in~\cite{Ge2016}, we propose a deeper network based on the stacked hourglass network that currently gives state-of-the-art results for human pose tracking and human face landmark detection~\cite{newell2016stacked,bulat2017far,yang2017stacked,bulat2016two}.

The inspiration for this work comes from recent advances in human pose tracking, where very deep convolutional networks have been trained to identify feature points on humans in a variety of poses and environments~\cite{newell2016stacked}. The method is based on heatmap regression, where each individual landmark is coded as a heatmap. The idea behind the used {\em hourglass network} is based on an  aggregation of local evidence~\cite{bulat2016convolutional,bulat2017far}, where heatmaps of individual landmarks are created in the first part of the network and fed into the next layer, thus enabling the network to refine its knowledge of the spatial coherence of landmark patterns. A recent paper~\cite{Deng2018}, estimates 3D landmarks from 2D photos using the joint correspondence between frontal and profile landmarks and that method gets state-of-the-art results on standard test sets used for facial tasks. We propose a network architecture resembling the architecture described in~\cite{Deng2018} but our end goal is different.

In our work, the metric is the landmark localisation error measured in physical units (in this case, mm) since the end application is often related to physical estimation of facial morphology. However, in most state-of-the-art methods in landmark placement based on facial 2D photos, the accuracy is measured as Normalised Mean Error (NME), which is the
average of landmark errors normalised by the bounding box
size~\cite{Deng2018}, which is not a particular good measure when working with 3D surfaces. In the work of~\cite{bulat2017far,Deng2018}, 3D landmarks are estimated purely from 2D photos by regressing the unknown z-coordinate using a CNN network. The landmark accuracy in these cases is highly dependent on the heatmap resolution. They also do not utilise the availability of a true underlying 3D surface. In our work, we demonstrate a novel way to combine the output from state-of-the-art networks in a geometric based consensus to produce highly accurate 3D landmark predictions.




\section{Methods}

A summary of the method can be seen in Fig.~\ref{fig:CompleteFlow}. The overall idea is that the 3D surface is rendered from multiple views and for each view, landmark candidates are estimated. Each landmark estimate is now considered as a ray in 3D space and the final landmark 3D position is estimated using an outlier robust least squares estimate between {\em landmark rays}.

\subsection{Multi-view rendering}

The scans are rendered using an OpenGL rendering pipeline. The scan is placed approximately at the origin of the coordinate system and the camera is placed in 100 random positions around the face, and with the focal point at the origin. The parameters of the camera are determined so the entire scan is in view. We use an orthographic projection. For each view, we render a view of the scan and store the 2D coordinates of the projected ground-truth 3D landmarks. A simple ambient white light source is used. All surfaces are rendered in a monochrome non-texture setup and surfaces with texture are also rendered with full RGB texture colours.

It is an ill-posed problem to pre-align a general surface scan to a canonical direction as also described in~\cite{su2015multi}. However, for facial surfaces it is common to start the pipeline by identifying feature points such as the eyes and the nose and using them to pre-align the scan or to crop an area of interest as in~\cite{grewe2016fully}. In this work, we do not rely on pre-alignment and simply have the loose assumption that the scanned surface contains a face among other information, such as shoulders, and that an approximate direction of the face is given by the scanner. It also means that we do not assume that a given camera position can be in any fixed position with relation to the facial anatomy. This makes the algorithm general and not specific to facial anatomies. The rendering pipeline is used in both generating 2D training images and computing the landmarks on an unseen 3D surface.

\subsection{Geometric Derivatives}

To enable the implicit use of geometry, a set of geometric representations are also rendered. The first is a distance map representation that is computed as being the OpenGL z-buffer, where the precision has been optimised by setting the near and far clipping planes as close as possible to the scans bounding box. Using a depth map for human feature recognition was popularised in the seminal articles on pose recognition using depth sensors~\cite{shotton2011real} and later applied in, for example, 3D estimation of face geometry from 2D photos~\cite{sela2017unrestricted}. The standard geometry-only view is also rendered by disabling the texture. This is the representation used in the multi-view papers~\cite{qi2016volumetric,su2015multi}.

While curvature is implicitly represented in the depth map, we also render a view where the surface is grayscale coded according to the local mean curvature. Our aim is to use the method on surfaces containing surface noise, and therefore a robust curvature estimation is needed. Traditional methods that estimate curvature based on 1-ring neighbours of a vertex are too noisy. We use an approach where for each vertex $P$, the algorithm finds the curvature of the sphere that passes through $P$ and that best approximates the set of neighbours $P_N$ of $P$. The curvature is estimated by first doing an inverse projection of neighbouring points, so they are lying on an approximate plane, and then performing an eigenvector analysis of this point cloud. In this work, neighbour points $P_N$ are found by a mesh based region growing algorithm that includes points connected to $P$ at a maximal distance empirically chosen to be 10 mm. Proofs and implementation details can be found in~\cite{Delingette94Thesis,paulsen2004statistical}.


\subsection{Network Architecture and Loss Function}

\begin{figure}[htbp]
  \centering
  \includegraphics[width=.99\linewidth]{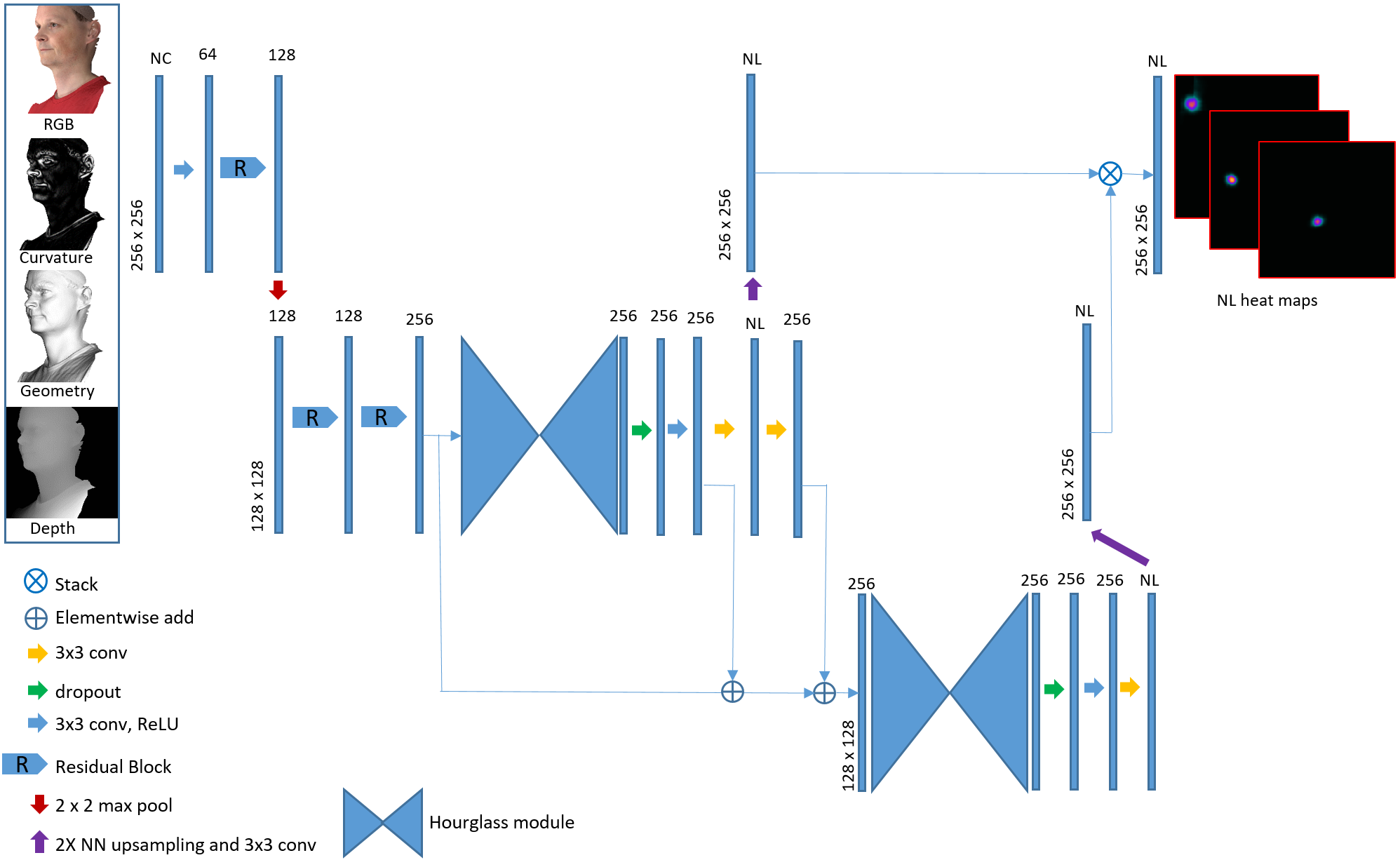}
  \caption{Network architecture. The input is an image with a varying number of channels (NC) and the output is one heatmap per landmark. NL is the number of landmarks. Blue boxes are feature maps and the number on top is the number of feature maps. The spatial size of the layer is written in the lower left corner (where necessary).}
  \label{fig:NetWorkArchitecture}
\end{figure}

We use the two-stack hourglass model described in~\cite{newell2016stacked}, which is based on the residual blocks described in~\cite{he2016deep}.  We focus on having a higher resolution of the predicted heatmap than in previous work. After the heatmap prediction in the hourglass block, the heatmap is upsampled twice using nearest neighbours followed by a 3x3 learnable convolutional layers as suggested in~\cite{odena2016deconvolution}. The input to the network for a single view is 2D renderings of the 3D surface. The network is flexible with the number of input layers. Using renderings of the textured surface (\texttt{RGB}) adds three layers, while the \texttt{depth} rendering, the \texttt{geometry} rendering and the \texttt{curvature} rendering each add one layer. The input images are 256 x 256, the dimensions throughout the hourglass stacks are 128 x 128 and the heatmap is upsampled to 256 x 256. The entire network can be seen in Fig.~\ref{fig:NetWorkArchitecture}. The ground truth is one heatmap per landmark, with a Gaussian kernel placed at the projected 2D position of the landmark. A cross entropy loss function is used. The heatmap estimates from the first and second hourglass modules are concatenated together to form a combined loss function, as demonstrated in previous work~\cite{newell2016stacked}. This ensures intermediate supervision of the network. Only one network is used and is able to recognise landmarks from all view directions.

The network is implemented with Tensorflow and trained using \texttt{RMSPROP} with an initial learning rate of 0.00025, a decay of 0.96 and a decay step of 2000. The batch size is between 4 and 8, the drop-out rate is 0.02. The network converged after around 60-100 epochs depending on the used rendering input. The network was trained and evaluated on one NVIDIA Titan X GPU card. The network is not optimised with respect to processing time and currently, it takes approximately 20 seconds to process 100 renderings, including overhead for reading and storing intermediate results.

\subsection{Landmark Detection and Consensus Estimation}

For a given input image, the output of the network are $NL$ heatmaps, where $NL$ is the number of landmarks. A 2D landmark is found as the position of the maximum value in the associated heatmap. This is illustrated in the top box of Fig.~\ref{fig:CompleteFlow}. However, to avoid using landmarks that are obviously not located correctly, only landmarks belonging to heatmap maxima over a certain threshold are considered. We have experimentally chosen a threshold of $0.5$. When a 2D landmark candidate has been found, the parameters of the corresponding 3D ray are computed using the inverse camera matrix used (and stored) for that rendering. Details of camera geometry and view rays can be found in~\cite{hartley2003multiple}. This is illustrated in the bottom part of  Fig.~\ref{fig:CompleteFlow} (3D view rays). For a given landmark, this results in up to $NV$ rays in 3D space, where $NV$ is the number of rendered views.


When the set of potential landmark rays for a given landmark has been computed, the landmark can be found as the crossing of these rays. In practice, the rays will not meet in a single point and some of the rays will be outliers due to incorrect 2D landmark detections. In order to robustly estimate a 3D point from several potentially noisy rays, we use a least squares (LSQ) fit combined with RANSAC selection~\cite{fischler1987random}. When each ray is defined by an origin $\mathbf{a}_i$ and a unit direction vector $\mathbf{n}_i$, the sum of squared distances from a point $\mathbf{p}$ to all rays is:

\begin{equation}\label{eq:SSDPoint}
  \sum_{i} \mathbf{d}^2_i = \sum_{i} \Big[  (\mathbf{p} - \mathbf{a}_i)^T (\mathbf{p} - \mathbf{a}_i) -
  \big[(\mathbf{p} - \mathbf{a}_i)^T \mathbf{n}_i \big]^2  \Big]
\end{equation}

Differentiating with respect to $\mathbf{p}$ results in a solution $\mathbf{p} = \mathbf{S}^{+} \mathbf{C}$, where $\mathbf{S}^{+}$ is the pseudo-inverse~\cite{ben2003generalized} of $\mathbf{S}$. Here $\mathbf{S} = \sum_{i} (\mathbf{n}_i \mathbf{n}^T_i - \mathbf{I})$ and $\mathbf{C} = \sum_{i} (\mathbf{n}_i \mathbf{n}^T_i - \mathbf{I}) \mathbf{a}_i$

In the RANSAC procedure, the initial estimate of $\mathbf{p}$ is based on three random rays and the residual is computed as the sum of squared distances from the included rays to the estimated $\mathbf{p}$. The result is a robust estimate of a 3D point based on a consensus between rays where outlier rays are not included (Fig.~\ref{fig:CompleteFlow}, LSQ+RANSAC). The final 3D landmark estimation is done by projecting the found point to the closest point on the target surface using an octree based space-division search algorithm. The result can be seen in Fig.~\ref{fig:CompleteFlow}(right).

\section{Data}

\begin{figure}[htbp]
  \centering
  \includegraphics[width=.40\linewidth]{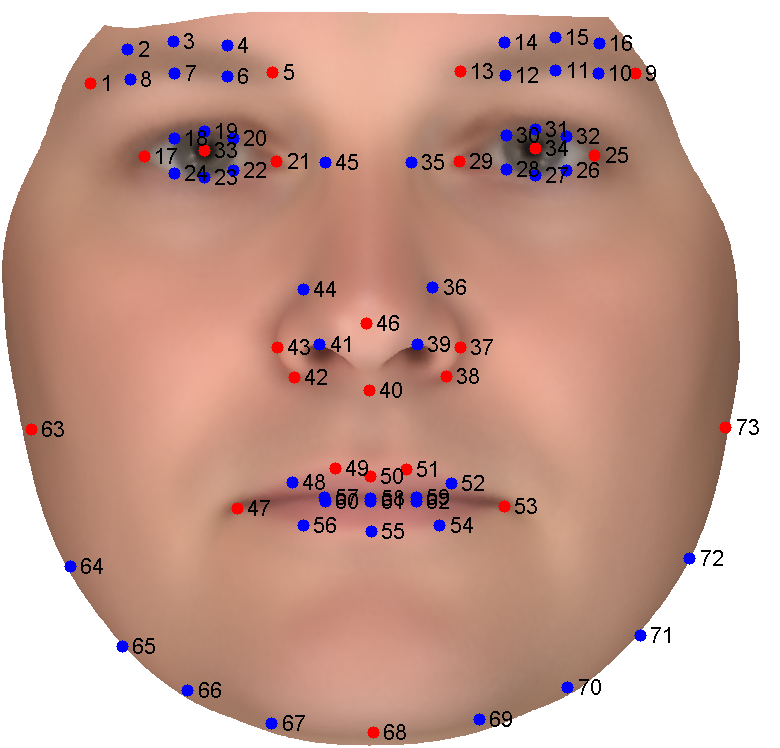}
  \includegraphics[width=.59\linewidth]{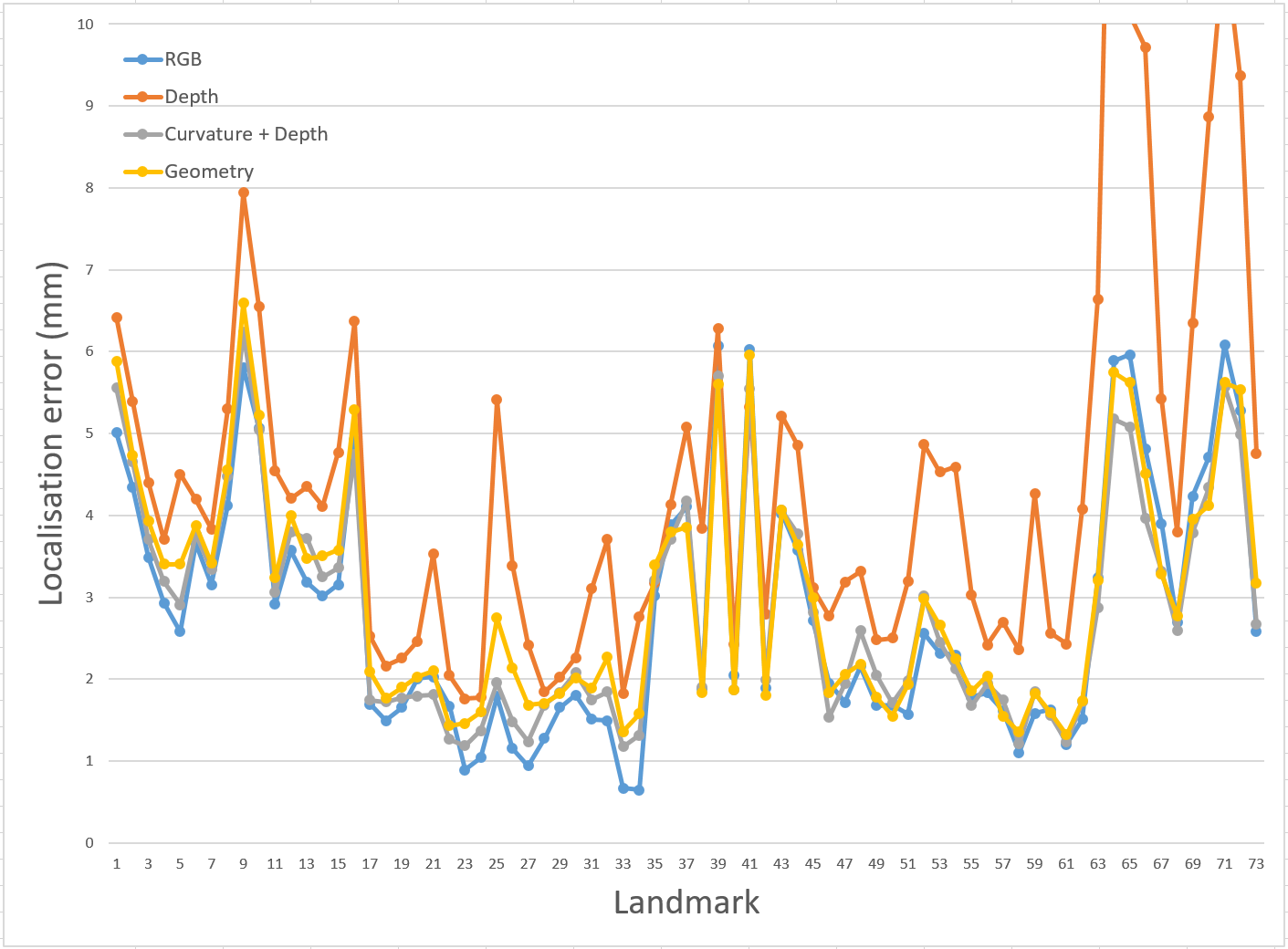}
  \caption{\textbf{Left)} The 73 landmark scheme used on the \textbf{DTU-3D} database~\cite{fagertun20143d}. \textbf{Right)} Landmark localisation errors on the \textbf{DTU-3D} database with different rendering configurations.}
  \label{fig:Face73LM}
\end{figure}

\paragraph{\textbf{DTU-3D} } The dataset consists of facial 3D scans of 601 subjects acquired using a \texttt{Canfield Vectra M3} surface scanner. The scanner accuracy is specified to be in the order of 1.2 mm (triangle edge length). Each face has been annotated with 73 landmarks using the scheme described in~\cite{fagertun20143d} and seen in Fig.~\ref{fig:Face73LM}. The faces in this dataset are all captured with a neutral expression. The data are used without being cropped, so both ears, neck and shoulders are partly present (see Fig.~\ref{fig:CompleteFlow} for an example scan from the database).

\paragraph{\textbf{BU-3DFE}} The database contains 100 subjects (56 female, 44 male). Each subject performed seven expressions in front of the 3D face scanner, resulting in 2,500 3D textured facial expression models in the database~\cite{yin20063d}. Each scan is annotated with 83 landmarks and the faces have been cropped to only contain the facial region.

\begin{figure}[htbp]
  \centering
  \includegraphics[width=.99\linewidth]{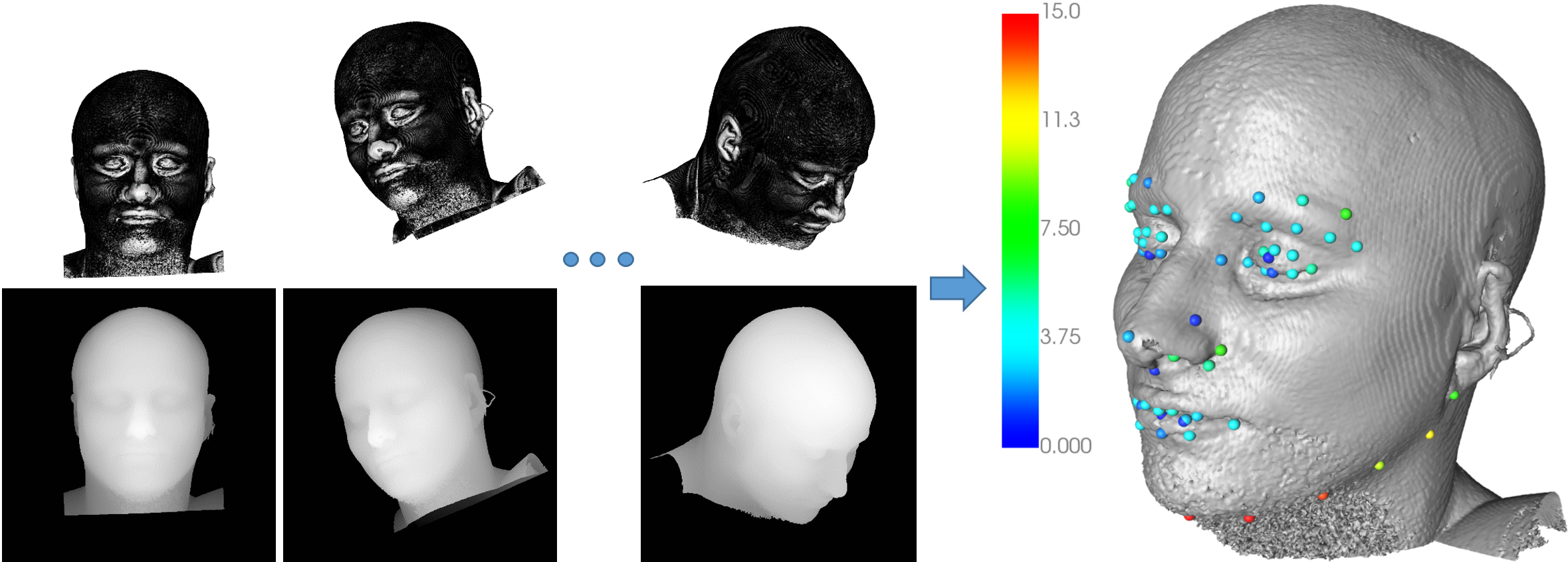}
  \caption{Finding landmarks on iso-surfaces extracted from an MR image using combined curvature and depth rendering. \textbf{Left)} examples of the rendered curvature and depth images. \textbf{Right)} the 3D landmarks identified on the skin iso-surface. The colour coding is the localisation error compared to manually annotated landmarks.}
  \label{fig:MRPipeline}
\end{figure}

\paragraph{\textbf{MR}} One Magnetic Resonance (MR) volume of a human head acquired using a 3T Siemens Trio scanner with a T1-weighted MEMPRAGE sequence with an isotropic voxel size of 1 mm. We use the N3 algorithm~\cite{sled1998nonparametric} for bias field correction and the intensity normalisation approach from~\cite{dale1999cortical}. The outer skin surface is extracted using marching cubes~\cite{lorensen1987marching} with an experimentally chosen iso-surface value of 20. The surface can be seen in Fig.~\ref{fig:MRPipeline}(right); the skin surface is noisy due to inconsistences in MR values around the skin-air interface.


\section{Results and Discussion}

The \textbf{DTU-3D} data were divided into a training set with 541 faces and a validation set of 60 faces. For both the training and the validation set, the faces were rendered from 100 camera positions using \texttt{RGB}, \texttt{geometry}, \texttt{curvature}, and \texttt{depth} rendering. The network is trained and validated using different render combinations (as seen in Fig.~\ref{fig:Face73LM}). The \textbf{BU-3DFE} was divided into a training set with all scans belonging to 46 females and 34 males and a validation set with all scans belonging to 10 females and 10 males. This results in a training set of 2000 faces and a validation set of 500 faces. The network using \textbf{BU-3DFE} was trained and evaluated using \texttt{RGB} renderings.

For a given validation face, the 3D landmarks are computed using the proposed method and compared with the ground truth landmarks. The landmark localisation error is computed as the Euclidean distance between an annotated landmark and an estimated landmark. We report the mean and standard deviation of the landmark localisation error per landmark in millimeters as seen in Tab.~\ref{tab:BU-3DFE_results} and Fig.~\ref{fig:Face73LM}.


\begin{table}
\begin{center}
\begin{tabular}{ | l | l | l | l | l | l | l | l | l | }
\hline
	         & \multicolumn{2}{|l|}{Salazar~\cite{salazar2014fully}} & \multicolumn{2}{|l|}{Gilani~\cite{gilani2015shape}} & \multicolumn{2}{|l|}{Grewe~\cite{grewe2016fully}} & \multicolumn{2}{|l|}{This paper} \\ \hline
    images & \multicolumn{2}{|l|}{350} & \multicolumn{2}{|l|}{2500} & \multicolumn{2}{|l|}{2500} & \multicolumn{2}{|l|}{500} \\ \hline
	      & Mean & impr. & Mean & impr. & Mean & impr. & Mean & SD \\ \hline
	Ex(L) (17) & 9.63 & 73\% & 4.43 & 41\% & 2.95 & 11\% & 2.59 & 1.53 \\ \hline
	En(L) (21) & 6.75 & 71\% & 4.75 & 64\% & 3.04 & 37\% & 1.89 & 0.98 \\ \hline
	Ex(R)(25) & 8.49 & 66\% & 4.34 & 33\% & 3.22 & 11\% & 2.85 & 1.5 \\ \hline
	En(R) (29) & 6.14 & 70\% & 3.29 & 33\% & 3.23 & 44\% & 1.8 & 0.89 \\ \hline
	Ac(L) (42) & 6.47 & 59\% & 4.3 & 38\% &  &  & 2.61 & 1.41 \\ \hline
	Ac(R ) (38) & 7.17 & 58\% & 4.28 & 29\% &  &  & 2.96 & 1.56 \\ \hline
	Sn (40)     &  &  & 3.9 & 31\% & 1.97 & -29\% & 2.52 & 1.69 \\ \hline
	Ch(L) (47) &  &  & 6 & 86\% &  &  & 2.182 & 1.44 \\ \hline
	Ch(R) (53) &  &  & 5.45 & 68\% &  &  & 2.42 & 1.44 \\ \hline
	Ls (50) &  &  & 3.2 & 19\% &  &  & 2.33 & 1.31 \\ \hline
	Li (55) &  &  & 6.9 & 99\% &  &  & 2.5 & 1.41 \\ \hline
\hline
	Mean & 7.44 &  & 4.62 &  & 2.88 &  & 2.42 &  \\ \hline
\end{tabular}
\end{center}
\caption{Results on the \textbf{BU-3DFE} database. Error is in mm. Improvements compared to previous work are in percentages. The number in parentheses after the landmark id is the corresponding landmark on Fig.~\ref{fig:Face73LM}.}\label{tab:BU-3DFE_results}
\end{table}



As can be seen in Fig.~\ref{fig:Face73LM}, the method locates a large set of landmarks with a localisation error in the range of 2 mm. This is in the limit of what an experienced operator can achieve~\cite{fagertun20143d}. The landmarks that have high errors are also the landmarks that are typically very difficult for a human to place. Landmarks placed on the chin and on the eyebrows are very hard to place manually, due to the weak anatomical cues. An extensive analysis of landmark errors can be found in~\cite{fagertun20143d}, where the inter-observer variance is reported for the landmark set used in the \textbf{DTU-3D}. It was found that the landmark around the chin has a manual inter-observer in the range of $6 mm$. The errors in Fig.~\ref{fig:Face73LM} for the \textbf{DTU-3D} base is probably more due to inconsistency in the manual annotation than from the presented method. The method handles uncropped 3D scans where both partial hair, ears, neck and shoulders are present as seen in Fig.~\ref{fig:CompleteFlow}, meaning that the only pre-processing step needed is a rough estimate of the overall direction of the head.

In Fig.~\ref{fig:Face73LM} it can be seen that rendering using RGB textures generally performs very well. This is not surprising since textured surfaces contain many visual cues. The mode where depth is used in combination with the curvature also performs very well. This result enables the method to be used on data where RGB texture is not naturally present, such as iso-surfaces from modalities like computed tomography (CT) or MR imaging. Using the depth layer alone yields reasonable results on many landmarks, but in particular around the chin, where few visual cues are seen in the depth image, the errors are large. Using \texttt{geometry} rendering yields slightly worse results than the \texttt{RGB} and \texttt{depth+curvature} renderings, but still on par with the state-of-the-art algorithms. We also tested the \texttt{RGB + depth + curvature + geometry} configuration but did not achieve superior results.

The generalisability of the method was tested on the \textbf{MR} scan using \texttt{depth + curvature} rendering. The method successfully locates landmarks on the MR iso-surface despite significant noise. The landmarks around the eyes and nose have an error level in the range of 2-3 mm, while the errors on the chin are larger, as also seen in the results on the pure range scan. It can be seen that the estimation of the curvature yields reasonable visual results despite the very high noise levels on the surface. Placing landmarks in 3D volumes is  notoriously difficult. While deep learning approaches are being used for true 3D landmark localisation~\cite{zheng20153d}, the methods are still limited by the prohibitive GPU memory requirements for true 3D processing. We believe that our approach offers an alternative way to handle complicated landmark placement in 3D volumes.

We have experimented with the number of views, using 25, 50, 75, and 100 view renderings. Going from 25 to 100 views decreases the landmark localisation error with less than a millimeter, meaning that in future applications the number of renderings could be significantly lowered. We have chosen to use random camera positions instead of selecting a fixed set of pre-defined positions. We believe that using random positions works as an extra data augmentation step. Finally, the results on the benchmark data set \textbf{BU-3DFE} as seen in Tab.~\ref{tab:BU-3DFE_results} demonstrate that the proposed method outperforms state-of-the-art methods by a large margin.

Compared to the methods that are dominant for 2D photos faces-in-the-wild as for example~\cite{bulat2017far,Deng2018,bulat2016two}, our metric is a landmark localisation error in mm, while they report the error as given as a percentage compared to either the eye-to-eye distance~\cite{bulat2016two} or the bounding box diagonal length~\cite{bulat2017far,Deng2018}. The average adult eye-to-eye distance is 62 mm. In~\cite{bulat2016two} they report an error rate of $4.5\%$ that is equivalent to 2.8 mm. This is a very impressive result but their measured error is still above what we report.

Since the 3D geometry of the face is available, it is possible to do a view-ray versus surface intersection test to for example determine if a landmark is placed on an occluded part of the surface. We have chosen not do that, since the method in practise is good at predicting the positions of landmarks on occluded parts of the face due to the high spatial correlation of landmark positions as for also demonstrated in~\cite{bulat2017far,Deng2018,bulat2016two}. Furthermore, the landmarks with low confidence are removed in the RANSAC procedure and this is often landmarks that from a given viewpoint are hard to see due to for example occlusion.

Future works includes improved maxima selection in the heatmap based on more rigorous statistical assumptions and using a fitting function to locate maxima with sub-pixel precision. In addition, different renderings and surface signature techniques to further enhance surface structure can be exploited. Furthermore, with increased GPU memory in sight, it will be interesting to design novel network architectures with higher spatial resolutions in mind and also test other blocks like the inception-resnet blocks used in~\cite{Deng2018}.



\section{Conclusion}

In this paper, we demonstrated a  multi-view pipeline to accurately locate 3D landmarks on facial surfaces. The method outperforms previous methods by a large margin with respect to landmark localisation error. Furthermore, the effect of using different rendering types was demonstrated, suggesting that a model trained on high-resolution 3D face scans could be used directly to accurately predict landmarks on surfaces from a completely different scanning modality. It was demonstrated that using a combination of surface curvature rendering and depth maps performs on par with using RGB texture rendering, proving that implicit geometric information can be very valuable even when observed in 2D.

While geometric deep learning is a rapidly growing field and volumetric methods are gaining foothold, this paper shows that the concept of multi-view rendering of 3D surfaces currently produces state-of-the-art results with regard to feature point location.

Code, trained models and test data can be found here:\\ \texttt{http://ShapeML.compute.dtu.dk/}

\section*{Acknowledgements}
We gratefully acknowledge the support of NVIDIA Corporation with the donation of the Titan Xp GPU used for this research.

\bibliographystyle{splncs04}
\bibliography{ACCV2018-RAPA}

\begin{thebibliography}{10}
\providecommand{\url}[1]{\texttt{#1}}
\providecommand{\urlprefix}{URL }
\providecommand{\doi}[1]{https://doi.org/#1}

\bibitem{ben2003generalized}
Ben-Israel, A., Greville, T.N.: Generalized inverses: theory and applications.
  Springer (2003)

\bibitem{blanz1999morphable}
Blanz, V., Vetter, T.: A morphable model for the synthesis of 3d faces. In:
  Proc. Computer graphics and interactive techniques. pp. 187--194 (1999)

\bibitem{boscaini2016learning}
Boscaini, D., Masci, J., Rodol{\`a}, E., Bronstein, M.: Learning shape
  correspondence with anisotropic convolutional neural networks. In: Proc.
  NIPS. pp. 3189--3197 (2016)

\bibitem{bowyer2006survey}
Bowyer, K.W., Chang, K., Flynn, P.: A survey of approaches and challenges in 3d
  and multi-modal 3d+ 2d face recognition. Computer vision and image
  understanding  \textbf{101}(1),  1--15 (2006)

\bibitem{bronstein2017geometric}
Bronstein, M.M., Bruna, J., LeCun, Y., Szlam, A., Vandergheynst, P.: Geometric
  deep learning: going beyond {E}uclidean data. IEEE Signal Processing Magazine
   \textbf{34}(4),  18--42 (2017)

\bibitem{bulat2016convolutional}
Bulat, A., Tzimiropoulos, G.: Convolutional aggregation of local evidence for
  large pose face alignment. In: Proc. BMVC (2016)

\bibitem{bulat2016two}
Bulat, A., Tzimiropoulos, G.: Two-stage convolutional part heatmap regression
  for the 1st 3d face alignment in the wild (3dfaw) challenge. In: Proc. ECCV.
  pp. 616--624 (2016)

\bibitem{bulat2017far}
Bulat, A., Tzimiropoulos, G.: How far are we from solving the 2d \& 3d face
  alignment problem?(and a dataset of 230,000 3d facial landmarks). arXiv
  preprint arXiv:1703.07332  (2017)

\bibitem{chang2015three}
Chang, J.B., Small, K.H., Choi, M., Karp, N.S.: Three-dimensional surface
  imaging in plastic surgery: foundation, practical applications, and beyond.
  Plastic and reconstructive surgery  \textbf{135}(5),  1295--1304 (2015)

\bibitem{creusot2013machine}
Creusot, C., Pears, N., Austin, J.: A machine-learning approach to keypoint
  detection and landmarking on 3d meshes. International journal of computer
  vision  \textbf{102}(1-3),  146--179 (2013)

\bibitem{dale1999cortical}
Dale, A.M., Fischl, B., Sereno, M.I.: Cortical surface-based analysis: I.
  segmentation and surface reconstruction. Neuroimage  \textbf{9}(2),  179--194
  (1999)

\bibitem{Delingette94Thesis}
Delingette, H.: Mod\'{e}lisation, D\'{e}formation et Reconnaissance d'Objets
  Tridimensionnels \'{a} l'Aide de Maillages Simplexes. Ph.D. thesis,
  L'\'{E}cole Centrale de Paris (1994)

\bibitem{fagertun20143d}
Fagertun, J., Harder, S., Rosengren, A., Moeller, C., Werge, T., Paulsen, R.R.,
  Hansen, T.F.: 3d facial landmarks: Inter-operator variability of manual
  annotation. BMC medical imaging  \textbf{14}(1), ~35 (2014)

\bibitem{fischler1987random}
Fischler, M.A., Bolles, R.C.: Random sample consensus: a paradigm for model
  fitting with applications to image analysis and automated cartography. In:
  Readings in computer vision, pp. 726--740. Elsevier (1987)

\bibitem{Ge2016}
Ge, L., Liang, H., Yuan, J., Thalmann, D.: Robust 3d hand pose estimation in
  single depth images: From single-view cnn to multi-view cnns. In: Proc. CVPR.
  pp. 3593--3601 (2016)

\bibitem{gilani2015shape}
Gilani, S.Z., Shafait, F., Mian, A.: Shape-based automatic detection of a large
  number of 3d facial landmarks. In: Proc. CVPR. pp. 4639--4648. IEEE (2015)

\bibitem{goodfellow2016deep}
Goodfellow, I., Bengio, Y., Courville, A., Bengio, Y.: Deep learning. MIT press
  (2016)

\bibitem{gordon1992face}
Gordon, G.G.: Face recognition based on depth and curvature features. In:
  Proceedings 1992 IEEE Computer Society Conference on Computer Vision and
  Pattern Recognition. pp. 808--810. IEEE (1992)

\bibitem{grewe2016fully}
Grewe, C.M., Zachow, S.: Fully automated and highly accurate dense
  correspondence for facial surfaces. In: Proc. ECCV. pp. 552--568. Springer
  (2016)

\bibitem{hammond20043dshort}
Hammond, P., et~al.: 3d analysis of facial morphology. American journal of
  medical genetics Part A  \textbf{126}(4),  339--348 (2004)

\bibitem{hartley2003multiple}
Hartley, R., Zisserman, A.: Multiple view geometry in computer vision.
  Cambridge university press (2003)

\bibitem{he2016deep}
He, K., Zhang, X., Ren, S., Sun, J.: Deep residual learning for image
  recognition. In: Proc. CVPR. pp. 770--778 (2016)

\bibitem{Deng2018}
J.~Deng, Y.~Zhou, S.C., Zafeiriou, S.: Cascade multi-view hourglass model for
  robust 3d face alignment. In: FG (2018)

\bibitem{jourabloo2015pose}
Jourabloo, A., Liu, X.: Pose-invariant 3d face alignment. In: Proc. ICCV. pp.
  3694--3702 (2015)

\bibitem{lorensen1987marching}
Lorensen, W.E., Cline, H.E.: Marching cubes: A high resolution 3d surface
  construction algorithm. In: ACM siggraph computer graphics. vol.~21, pp.
  163--169. ACM (1987)

\bibitem{newell2016stacked}
Newell, A., Yang, K., Deng, J.: Stacked hourglass networks for human pose
  estimation. In: Proc. ECCV. pp. 483--499. Springer (2016)

\bibitem{odena2016deconvolution}
Odena, A., Dumoulin, V., Olah, C.: Deconvolution and checkerboard artifacts.
  Distill  (2016). \doi{10.23915/distill.00003}

\bibitem{paulsen2004statistical}
Paulsen, R.R.: Statistical shape analysis of the human ear canal with
  application to in-the-ear hearing aid design. Ph.D. thesis, Technical
  University of Denmark (2004)

\bibitem{paulsen2017creating}
Paulsen, R.R., Marstal, K.K., Laugesen, S., Harder, S.: Creating ultra dense
  point correspondence over the entire human head. In: Proc. SCIA. pp.
  438--447. Springer (2017)

\bibitem{perakis20133d}
Perakis, P., Passalis, G., Theoharis, T., Kakadiaris, I.A.: 3d facial landmark
  detection under large yaw and expression variations. IEEE transactions on
  pattern analysis and machine intelligence  \textbf{35}(7),  1552--1564 (2013)

\bibitem{qi2016volumetric}
Qi, C.R., Su, H., Nie{\ss}ner, M., Dai, A., Yan, M., Guibas, L.J.: Volumetric
  and multi-view cnns for object classification on 3d data. In: Proc. CVPR. pp.
  5648--5656 (2016)

\bibitem{salazar2014fully}
Salazar, A., Wuhrer, S., Shu, C., Prieto, F.: Fully automatic
  expression-invariant face correspondence. Machine Vision and Applications
  \textbf{25}(4),  859--879 (2014)

\bibitem{sedaghat2017orientation}
Sedaghat, N., Zolfaghari, M., Amiri, E., Brox, T.: Orientation-boosted voxel
  nets for 3d object recognition. In: British Machine Vision Conference (BMVC)
  (2017)

\bibitem{sela2017unrestricted}
Sela, M., Richardson, E., Kimmel, R.: Unrestricted facial geometry
  reconstruction using image-to-image translation. arXiv  (2017)

\bibitem{shotton2011real}
Shotton, J., Fitzgibbon, A., Cook, M., Sharp, T., Finocchio, M., Moore, R.,
  Kipman, A., Blake, A.: Real-time human pose recognition in parts from single
  depth images. In: Proc. CVPR. pp. 1297--1304 (2011)

\bibitem{sled1998nonparametric}
Sled, J.G., Zijdenbos, A.P., Evans, A.C.: A nonparametric method for automatic
  correction of intensity nonuniformity in mri data. IEEE transactions on
  medical imaging  \textbf{17}(1),  87--97 (1998)

\bibitem{su2015multi}
Su, H., Maji, S., Kalogerakis, E., Learned-Miller, E.: Multi-view convolutional
  neural networks for 3d shape recognition. In: Proceedings of the IEEE
  international conference on computer vision. pp. 945--953 (2015)

\bibitem{wiles2017silnet}
Wiles, O., Zisserman, A.: Silnet: Single-and multi-view reconstruction by
  learning from silhouettes. In: Proc. BMVC (2017)

\bibitem{yang2017stacked}
Yang, J., Liu, Q., Zhang, K.: Stacked hourglass network for robust facial
  landmark localisation. In: Proc. CVPR. pp. 2025--2033. IEEE (2017)

\bibitem{yin20063d}
Yin, L., Wei, X., Sun, Y., Wang, J., Rosato, M.J.: A 3d facial expression
  database for facial behavior research. In: Proc. FGR. pp. 211--216. IEEE
  (2006)

\bibitem{zheng20153d}
Zheng, Y., Liu, D., Georgescu, B., Nguyen, H., Comaniciu, D.: 3d deep learning
  for efficient and robust landmark detection in volumetric data. In: Proc.
  MICCAI. pp. 565--572. Springer (2015)

\bibitem{zhu2016face}
Zhu, X., Lei, Z., Liu, X., Shi, H., Li, S.Z.: Face alignment across large
  poses: A 3d solution. In: Proc. CVPR. pp. 146--155 (2016)

\end{thebibliography}

\end{document}